\documentclass[letterpaper]{article} 
\usepackage[preprint]{aaai2027}  
\usepackage[hyphens]{url}  
\usepackage{graphicx} 
\usepackage{natbib}  
\usepackage{caption} 
\usepackage{algorithm}
\usepackage{algorithmic}

\usepackage{booktabs}
\usepackage{amssymb}
\usepackage{amsmath}

\usepackage{newfloat}
\usepackage{listings}
\DeclareCaptionStyle{ruled}{labelfont=normalfont,labelsep=colon,strut=off} 
\floatstyle{ruled}
\newfloat{listing}{tb}{lst}{}
\floatname{listing}{Listing}

\title{PACMI: Provenance-Aware Cascading Memory Invalidation \\ for Long-Term LLM Agents}
\author{
    \large\mdseries 
    \mbox{%
    Yiqi Wang\textsuperscript{\rm 1,*},
    Jiaqi Liu\textsuperscript{\rm 2},
    Jiaqi Zhang\textsuperscript{\rm 3},
    Zhangkai Wu\textsuperscript{\rm 4},
    Yiqun Duan\textsuperscript{\rm 5},
    Mingkai Zheng\textsuperscript{\rm 2},
    Taotao Cai\textsuperscript{\rm 1}%
    }
}

\affiliations{
    \textsuperscript{\rm 1}University of Southern Queensland,
    \textsuperscript{\rm 2}Southern University of Science and Technology\\
    \textsuperscript{\rm 3}Jiangsu University,
    \textsuperscript{\rm 4}The University of Sydney,
    \textsuperscript{\rm 5}Uploading Inc.\\
    \textsuperscript{*}yiqi.wang.jennie@gmail.com
}
\begin{document}

\maketitle

\begin{abstract}
LLM agents rely on long-term memory to retain and reuse information when performing tasks over long horizons. Existing methods provide limited support for handling memories that become outdated as new observations or domain evidence arrive. Such outdated memories may remain semantically relevant, continue to affect dependent records, and retain value as historical evidence.
This calls for two capabilities: \emph{dependency tracking} to identify downstream effects and \emph{historical preservation} to retain useful past records. We propose \textbf{P}rovenance-\textbf{A}ware \textbf{C}ascading \textbf{M}emory \textbf{I}nvalidation (\textbf{PACMI}), a framework that represents memories and new evidence in a provenance graph with typed dependency edges. PACMI assigns records to a four-state validity lattice, propagates validity changes to dependent memories, and uses the resulting states for retrieval and stale-premise detection. We also introduce a diagnostic benchmark with 100 cases and 300 queries across five domains. The evaluation separates node-, context-, and answer-level performance. PACMI achieves the highest final-answer accuracy on this benchmark, and its paired difference from the strongest baseline is significant under an exact McNemar test. The premise checker achieves perfect precision, recall, and $F_1$ on the controlled query distribution. Cascading propagation primarily improves memory-state correctness: removing it increases final-answer errors from 3 to 11, but the paired difference does not reach the 0.05 significance threshold. Code and data will be made publicly available.
\end{abstract}

\section{Introduction}
\label{sec:intro}

\begin{figure*}[t]
\centering
\includegraphics[width=0.92\textwidth]{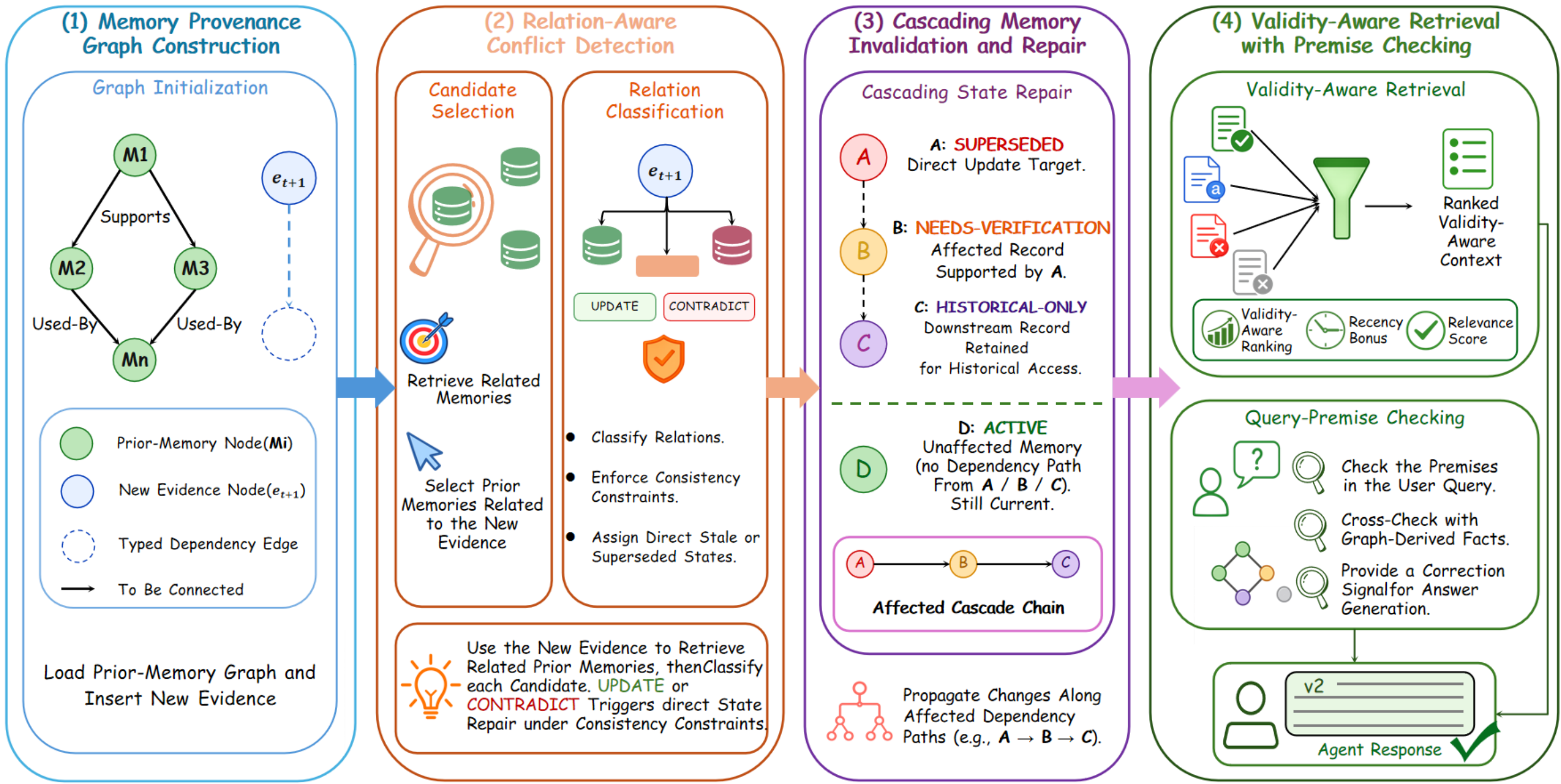}
\caption{Overview of PACMI. Starting from an annotated pre-update provenance graph, PACMI inserts new evidence, classifies evidence--memory relations, propagates validity changes along typed dependency edges, performs validity-aware retrieval, and checks whether the query assumes a superseded fact before response generation.}
\label{fig:pacmi-overview}
\end{figure*}

LLM agents~\cite{zhang2026the} can employ external tools to complete tasks progressively in a long-horizon. Through autonomous decision-making process, they have been widely applied to personalized assistance~\cite{park2023generativeagents}, task automation~\cite{yao2023react}, and multi-step problem solving~\cite{shinn2023reflexion,wang2024surveyagents}. These agentic tasks often require LLM agents to retain and reuse information across sessions rather than solve each task from scratch. Long-term memory has therefore become an important component of agent systems, supporting persistent user information~\cite{zhong2023memorybank,chhikara2025mem0, LEADer, yao2026agentarchitectureexecutionassumptions}, past interactions and task outcomes~\cite{packer2023memgpt,wang2023longmem, lu2025out, lu2026mllms,  lu2025deepresearch}, and accumulated experience for future tasks.

Equipped with long-term memory, existing memory-augmented agents are mainly designed to improve memory representation and information utilization. Text-based and vector-indexed methods store past information as retrievable records and retrieve or summarize relevant content when needed~\cite{lewis2020rag,chhikara2025mem0}. Temporal methods organize memories over time to preserve evolving information~\cite{rasmussen2025zep}. Graph-based methods link related entities, facts, events, or experiences to support structured access and multi-hop retrieval~\cite{edge2024graphrag,guo2024lightrag,gutierrez2024hipporag}. Recent systems further connect related notes and evolve memory structures to improve consolidation and long-term reuse~\cite{xu2025amem,lv2026allmem}. Although these methods improve memory storage, organization, and utilization, they pay limited attention to whether stored memories remain \emph{valid} after new evidence arrives.

\emph{Agent memory changes over time}. New observations may update a user's location, preference, goal, or plan. \emph{New domain evidence} may also conflict with previously stored information. However, LLM agents often fail to identify such invalid memories, especially under implicit updates~\cite{chao2026stale,tao2026memconflict}. Repeated LLM-based rewriting may further introduce errors~\cite{zhang2026faultymemories, yao2026measuringllmtutorsteach}. Semantic relevance alone cannot solve this problem, because a stale memory may remain highly relevant and still be retrieved.

\emph{Stale memories are not always useless}. They may no longer support current decisions, but may remain valid as historical records. They may also affect downstream claims, assumptions, or recommendations. For example, if a user moves from Sydney to San Francisco, the Sydney record should not support current local recommendations. Related memories, such as restaurant suggestions or commuting assumptions, should also be re-evaluated. However, they may still be useful for historical queries. Stale-memory management therefore requires both dependency tracking and historical preservation.


We propose \textbf{PACMI}, which represents prior memories and new evidence as nodes in a memory provenance graph with typed dependency edges. Given new evidence, PACMI classifies evidence--memory relations, identifies directly replaced records, propagates validity along dependency edges into a four-state lattice, and then performs validity-aware retrieval and stale-premise checking before response generation. Figure~\ref{fig:pacmi-overview} shows the workflow. Our main contributions are summarized as follows:

\begin{itemize}
\item We propose \textbf{PACMI}, a provenance-aware framework for stale-memory management. It uses a four-state validity lattice and typed-edge propagation to update the validity of dependent memories.

\item We release a \textbf{new diagnostic benchmark} with 100 cases and 300 queries across five domains. Its evaluation separates node-, context-, and answer-level performance.

\item PACMI achieves \textbf{higher final-answer accuracy} on the benchmark, with a significant paired difference from the strongest baseline. The premise checker is perfectly accurate on the controlled query-type evaluation, while cascading propagation improves memory-state correctness but does not produce a statistically significant final-answer difference at the evaluated scale.

\end{itemize}

\paragraph{Scope and assumptions.}
We state up front what this paper does and does not establish, because it determines how our numbers should be read.
\textbf{(A1)} The dependency edges of the pre-update graph are \emph{given} by the benchmark annotation. We study propagation and retrieval \emph{given} a provenance graph; we do not study graph induction from raw memory text.
\textbf{(A2)} Each case presents a single new-evidence item over a fixed two-hop dependency chain and six prior memories.
\textbf{(A3)} Consequently, in this benchmark the ground-truth validity labels are a deterministic function of graph topology. Metrics that score \emph{downstream} state assignment therefore verify that propagation is implemented correctly; they are not evidence of comparative advantage, and we do not present them as such (see the Experiments section).
These assumptions make the benchmark diagnostic rather than naturalistic. We argue this is the right first step for a problem that no existing benchmark isolates, and we specify in the Limitations section exactly which experiments would remove each assumption.

\section{Related Work}
\label{sec:related}

\paragraph{Long-term memory for LLM agents.}
Generative Agents maintain a memory stream of observations and reflections for long-horizon behavior \cite{park2023generativeagents}; Reflexion stores verbal feedback as episodic memory \cite{shinn2023reflexion}; MemGPT treats memory management as virtual context with explicit tiers \cite{packer2023memgpt}; MemoryBank and LongMem explore persistent user memory and long-context reuse \cite{zhong2023memorybank,wang2023longmem}; Mem0 extracts, consolidates, and retrieves salient memories at scale \cite{chhikara2025mem0}. These systems target storage, consolidation, and retrieval. PACMI instead targets whether stored memories remain valid once new evidence arrives.

\paragraph{Graph-structured memory, stale memory, and evaluation.}
GraphRAG and LightRAG show that graph indexing captures dependencies that flat chunks cannot \cite{edge2024graphrag,guo2024lightrag}; HippoRAG supports associative multi-hop retrieval \cite{gutierrez2024hipporag}; Zep uses a temporal knowledge graph for agent memory \cite{rasmussen2025zep}; A-MEM links new memories to related historical notes \cite{xu2025amem}; All-Mem maintains a topology-structured bank with non-destructive evolution \cite{lv2026allmem}. These improve organization and retrieval but do not track validity under evolving evidence. In parallel, STALE evaluates whether agents recognize invalid memories, including state resolution and premise resistance \cite{chao2026stale}, and MemConflict defines memory validity as query-conditioned fitness-for-use \cite{tao2026memconflict}; both focus on \emph{direct} conflicts between old and new information. Broader benchmarks---LoCoMo \cite{maharana2024locomo}, LongMemEval \cite{wu2024longmemeval}, and MemoryAgentBench \cite{hu2025memoryagentbench}---cover recall, temporal reasoning, updates, forgetting, and conflict resolution, but none asks whether a system can identify the \emph{downstream} memories affected by stale evidence, or distinguish current, historical, and invalidated use of the same record.

\paragraph{Dependency-directed belief revision.}
Tracking which conclusions depend on which premises, and retracting them when a premise fails, is a classical AI problem. Truth maintenance systems record justifications and perform dependency-directed retraction: a JTMS relabels dependent nodes when a justification becomes unsupported \cite{doyle1979tms}, and an ATMS maintains labels over assumption sets so that multiple contexts coexist \cite{dekleer1986atms}. AGM belief revision axiomatizes the accompanying principle of minimal change \cite{agm1985}. The database analogue is provenance and incremental view maintenance, where a base-tuple change invalidates exactly the derived views that depend on it \cite{gupta1995views,cheney2009provenance}, with PROV-DM providing a standard vocabulary for such dependencies \cite{moreau2013prov}.

PACMI is, in this lineage, a truth maintenance system over natural-language agent memory. Three differences drive the design. First, the relation between new evidence and candidate memories is expressed in natural language and must be inferred by an LLM under deterministic constraints. In the current benchmark, however, the pre-update dependency edges are supplied by annotation under assumption A1; inducing these edges from raw memory text remains
future work. Second, labels are not binary IN/OUT: an agent must rank rather than admit or reject, so validity is a graded state coupled to retrieval scoring. Third, retraction is not deletion: a superseded record remains the correct answer to an explicitly historical query, so invalidation must be non-destructive. To our knowledge this connection has not been made explicit in the agent-memory literature, and we make it here both to position the contribution and to indicate that the symbolic-TMS setting supplies a natural upper bound for future work.

\section{Problem Formulation}
\label{sec:problem}

Given newly observed evidence, the goal is not only to retrieve relevant memories, but to determine which remain valid, which are replaced, and which downstream memories or decisions are affected.

\paragraph{Memory provenance graph.}
Let $\mathcal{M}_t=\{m_1,m_2,\ldots,m_n\}$ denote the memories held before time step $t$, each with textual content, timestamp, source, and optional metadata. Rather than treating memories as independent records, we represent agent memory as a provenance graph $G_t=(V_t,E_t)$, where $V_t$ contains prior-memory and new-evidence nodes and $E_t$ contains typed dependency edges. Prior-memory nodes store textual records that may describe facts, plans, recommendations, claims, historical notes, or previous decisions; these semantic roles are annotations, not runtime node classes. The evaluated pre-update graph contains \textit{supports} and \textit{used-by} edges; after relation classification PACMI may add an \textit{updates} or \textit{contradicts} edge from the new evidence to an accepted direct target.

\paragraph{Task.}
At step $t+1$ the agent receives evidence $e_{t+1}$ and query $q$, and must produce $(G_{t+1},S_{t+1},C_q)$: the updated graph, a validity-state assignment over memory nodes, and the retrieved context used for generation. This decomposes into three sub-goals: identify records directly replaced by or in conflict with the new evidence; trace dependency edges to the downstream records affected by them; and construct a context $C_q$ that prioritizes appropriate evidence, reducing stale reuse while retaining history when the query is explicitly historical.

\paragraph{Validity lattice.}
We use four validity states, ordered from weaker to stronger non-current status; from left to right, their eligibility for \emph{current-oriented} retrieval decreases:
\[
\begin{aligned}
\textsc{active}\;&<\;\textsc{needs-verification}\\
&<\;\textsc{historical-only}\;<\;\textsc{superseded}.
\end{aligned}
\]
\textsc{Active} records support current reasoning directly. \textsc{Superseded} records state an older value that newer evidence has replaced. \textsc{Needs-verification} records depend on changed evidence and are admissible but discounted. \textsc{Historical-only} records are retained primarily for explicitly historical queries. The ordering is not a convention: it is exactly the ordering of the query-dependent retrieval biases used at inference time ($+0.12$, $+0.04$, $-0.10$, $-0.22$ for current-oriented queries; see Appendix~\ref{app:impl}), and it induces the "keep the stronger state'' rule used when multiple propagation paths reach the same node. The state assignment is non-destructive throughout: original text is preserved, so a superseded record remains available as historical evidence.\footnote{The implementation additionally distinguishes a \textsc{stale} state for direct \textit{contradicts} targets and an \textsc{invalidated} state reached through \textit{derived-from} edges, which is excluded from retrieval entirely. Neither is instantiated by the benchmark reported here, so we omit them from the formulation rather than report untested machinery.}

\paragraph{Running example.}
Consider a hotel-booking scenario, following the implicit-conflict setting of STALE \cite{chao2026stale}. Evidence $e_1$ supports memory $m_1$, which supports claim $c_1$, which is used by decision $d_1$. The agent later observes $e_2$: Hotel X is sold out on January 10. The relation between $e_2$ and $m_1$ is classified as \textit{update}, so $m_1$ moves from \textsc{active} to \textsc{superseded}; propagation along \textit{supports} moves $c_1$ to \textsc{needs-verification}, and along \textit{used-by} moves $d_1$ to \textsc{historical-only}. The related hard negative $m_2$ stays \textsc{active} because it does not depend on the superseded record. Standard retrieval may still return $m_1$ and $c_1$ and produce an outdated recommendation; PACMI's validity-aware ranking instead prioritizes the new evidence and active memories.

\begin{figure*}[t]
    \centering
    \includegraphics[width=0.95\textwidth]{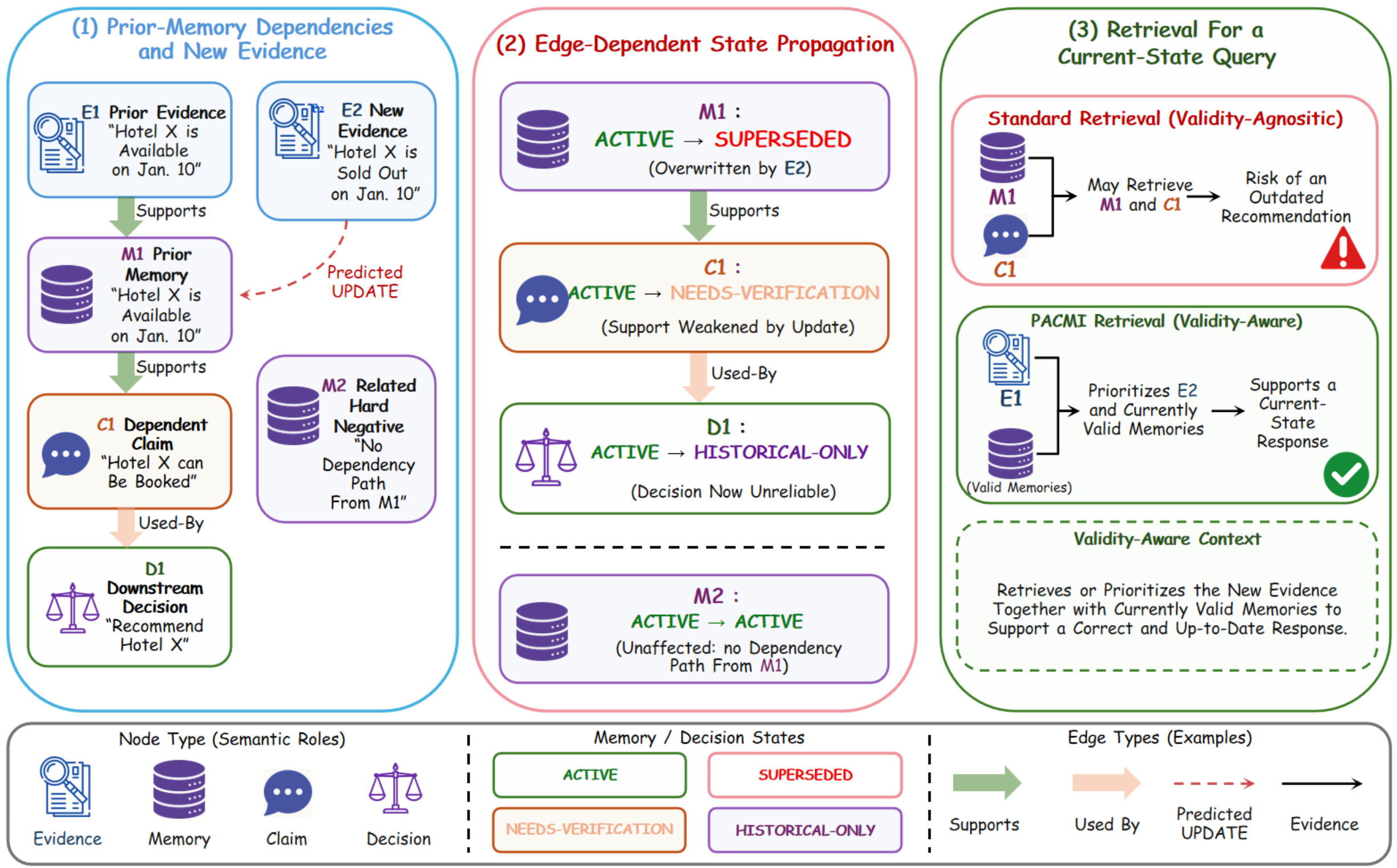}
    \caption{Running example of cascading memory invalidation. New evidence $e_2$ updates memory $m_1$, changing $m_1$ to \textsc{superseded}; propagation through \textit{supports} changes $c_1$ to \textsc{needs-verification}, and propagation through \textit{used-by} changes $d_1$ to \textsc{historical-only}, while the related hard negative $m_2$ remains \textsc{active}. Validity-aware retrieval then prioritizes the new evidence and active memories.}
    \label{fig:running-example}
\end{figure*}

Figure~\ref{fig:running-example} shows the resulting state assignment. Note that the hard negative $m_2$ is the case that distinguishes dependency-aware invalidation from similarity-based invalidation: it is close to $m_1$ in embedding space, so a system that invalidates by semantic proximity to the superseded record would demote it, whereas a system that invalidates along dependency edges leaves it \textsc{active}.

\section{Method}
\label{sec:method}

\paragraph{Overview.}
PACMI separates local semantic judgment from global validity propagation. It inserts new evidence, classifies evidence--memory relations with an auxiliary LLM under deterministic constraints, accepts directly replaced records, propagates edge-dependent states, and constructs a validity-aware context before generation.

\paragraph{Relation-aware conflict detection.}
Candidate ranking combines embedding similarity, token-level overlap, and matching semantic slots; exact weights are in Appendix~\ref{app:impl}. PACMI retains up to six candidates; because each evaluated case has six prior memories, ranking orders but does not prune candidates in this benchmark. For each candidate the auxiliary LLM predicts \textit{duplicate}, \textit{refine}, \textit{update}, \textit{contradict}, \textit{temporally-compatible}, or \textit{unrelated}, with a reason. Deterministic slot-, stability-, and temporality-aware constraints then normalize these predictions. Only \textit{update} and \textit{contradict} can create a direct target. A candidate with an incoming \textit{supports} or \textit{used-by} edge from another prior memory is treated as a downstream record rather than a direct target; its effect is handled by propagation. This split is what keeps local classification from having to reason about dependency structure.

\paragraph{Cascading invalidation and repair.}
From the accepted direct targets, PACMI performs breadth-first traversal over outgoing dependency edges. In the evaluated benchmark, each accepted direct target is an \textit{update} target and becomes \textsc{superseded}; the transition map is \textit{supports}$\rightarrow$\textsc{needs-verification} and \textit{used-by}$\rightarrow$\textsc{historical-only}. When several paths reach one node, the stronger state in the lattice wins. Unreached nodes remain \textsc{active}. Repair is non-destructive: PACMI records relation-judgment and cascade-invalidation operations while preserving the original text for traceability and historical retrieval.

\begin{algorithm}[t]
\caption{Provenance-Aware Cascading Memory Invalidation}
\label{alg:pacmi}
\small
\begin{algorithmic}[1]
\REQUIRE Pre-update memory graph $G_t$, new evidence $e_{t+1}$
\ENSURE Updated graph $G_{t+1}$ and inferred states $S_{t+1}$
\STATE Insert $e_{t+1}$; initialize all inference-time states as \textsc{active}
\STATE Rank up to $k$ prior memories to obtain candidates $\mathcal{C}$
\STATE $\mathcal{D}\leftarrow\emptyset$
\FOR{each $m_i\in\mathcal{C}$}
    \STATE Predict and constrain relation $r_i$ between $e_{t+1}$ and $m_i$
    \IF{$r_i$ is \textit{update}/\textit{contradict} and $m_i$ is a direct target}
        \STATE Add $m_i$ to $\mathcal{D}$; assign the relation-specific direct non-current state
        \STATE Add the predicted relation edge to $G_{t+1}$
    \ENDIF
\ENDFOR
\STATE Initialize queue $Q$ with $\mathcal{D}$
\WHILE{the queue is not empty}
    \STATE Remove $u$ from $Q$
    \FOR{each outgoing dependency edge $(u,v,\tau)$}
        \STATE Map $\tau$ to its downstream state $s_v$
        \STATE Strengthen $S_{t+1}(v)$ with $s_v$; enqueue newly affected $v$
    \ENDFOR
\ENDWHILE
\RETURN Updated graph $G_{t+1}$ and inferred state assignment $S_{t+1}$
\end{algorithmic}
\end{algorithm}

\paragraph{Validity-aware retrieval and premise checking.}
After propagation, PACMI ranks entries by semantic relevance, recency, slot match, new-evidence status, and query-dependent state biases. Current-oriented queries favor \textsc{active} and \textsc{needs-verification} records; historical queries instead favor \textsc{historical-only} and \textsc{superseded} records. The top five entries, rendered with state and slot labels, form $C_q$.

The premise checker consumes the \emph{inferred graph states}, not merely the retrieved context: it contrasts superseded prior records against the new evidence and active records. When it detects that a query assumes an outdated fact as current, PACMI adds the identified old and current facts to the answer prompt; otherwise it uses the same neutral prompt as the baselines. We report the premise checker's intrinsic conflict-classification performance separately from downstream final-answer accuracy. This intrinsic metric evaluates detection conditional on the inferred graph states; it does not by itself isolate the effect of the answer-side correction prompt.

\section{Experiments}
\label{sec:experiments}

We ask four questions. \textbf{RQ1:} how does PACMI compare with retrieval-based, revision-based, and graph-memory baselines on direct stale-node identification, context quality, stale reuse, and final answers? \textbf{RQ2:} does cascading invalidation identify downstream affected memories without over-invalidating valid information? \textbf{RQ3:} how do validity-aware retrieval, premise checking, and historical preservation affect final accuracy and resistance to stale premises? \textbf{RQ4:} how do individual components and query types affect performance?

\subsection{Benchmark}
\label{sec:benchmark}

Existing long-term memory benchmarks evaluate factual recall, temporal reasoning, or direct conflicts, with limited coverage of cases where new evidence invalidates not only a directly related memory but also downstream claims or decisions. We therefore construct a stale-memory dependency benchmark: 100 cases and 300 queries across five domains, each case with six prior memories, one new-evidence item, and three queries covering a current state, a stale premise, and a historical state. The evaluated dependency structure is the two-hop chain $M_1\xrightarrow{\textit{supports}}M_2\xrightarrow{\textit{used-by}}M_3$, with $M_4$--$M_6$ comprising a still-valid related memory, a semantically similar hard negative, and an unrelated distractor. Construction and annotation details are in Appendix~\ref{app:impl}.

\subsection{Baselines and Setup}
\label{sec:baselines}

\textbf{Vector RAG} uses semantic top-$k$ retrieval, and \textbf{Recency-RAG} adds a recency score. \textbf{LLM-Only Update} locally classifies and rewrites candidate memories before cosine retrieval. \textbf{CUPMem-style}, adapted from STALE \cite{chao2026stale}, supersedes records that directly state a replaced single-valued slot and admits previous values for historical queries, but does not propagate downstream effects. \textbf{A-MEM-style} constructs semantic links and expands linked notes without explicit validity propagation \cite{xu2025amem}. All methods use the same final-answer model, embedding model, and temperature-zero decoding; model, context-size, retrieval, and method-specific settings are in Appendix~\ref{app:impl}.

\subsection{Metrics}
\label{sec:metrics}

We evaluate node identification, retrieved context, and final answers separately; formal definitions are in Appendix~\ref{app:metrics}.
\emph{Direct Stale-Node Recall} (DR) and \emph{Affected-Node Recall} (AR) are case-macro recalls over the ground-truth direct and downstream sets. \emph{Over-} and \emph{Under-Invalidation} (OI, UI) are the fractions of cases whose predicted non-current set contains a node outside ground truth, or misses one inside it. \emph{Stale-in-Context} (SIC) is the fraction of retrieved entries that are stale or affected, and \emph{Current Valid-Evidence Recall} (CVER) is the fraction of active prior memories plus new evidence that is retrieved; both are averaged over current-oriented queries only, since superseded records may be appropriate evidence for a historical query. At the answer level, \emph{Final Accuracy} follows the annotated expected behavior, \emph{Stale Reuse} records whether an answer treats a directly stale memory as current, \emph{Premise Resistance} is accuracy in correcting the stale assumption in Q2, and Q3 accuracy operationalizes historical preservation. A fixed \texttt{deepseek-v4-pro} judge applies the same temperature-zero rubric to all methods.

\paragraph{Statistical analysis.}
We report percentile 95\% confidence intervals from 10,000 case-level bootstrap resamples with a fixed seed of 20260728. All three queries belonging to a sampled case are resampled together, while structural metrics retain their original case-level definitions. For prespecified paired binary comparisons, we use two-sided exact McNemar tests.

\paragraph{How to read the node-level metrics.}
By assumption A3, the ground-truth validity labels in this benchmark are a deterministic function of graph topology, and by A1 the topology is supplied as input. AR, UI, and recall-at-depth therefore certify that propagation is implemented correctly and that direct-target detection feeds it the right seeds; they do \emph{not} certify a comparative advantage over a method that also had the graph. We report them for completeness and for reproducibility of the pipeline, and we base no claim of superiority on them. The claims in this paper rest on the context-level and answer-level metrics, the prespecified paired significance tests, and the separately reported intrinsic premise-detector evaluation.

\subsection{Main Results}
\label{sec:main-results}

\begin{table*}[t]
\centering
\small
\setlength{\tabcolsep}{3pt}
\caption{Main results on the stale-memory dependency benchmark. Arrows give the preferred direction. ``--'' marks a metric the method cannot produce: Vector RAG, Recency-RAG, and A-MEM-style make no explicit invalidation decision, and only propagation-enabled methods predict downstream affected nodes. Per assumption A3, the AR/UI columns verify propagation correctness rather than comparative advantage. Point estimates are shown here; percentile 95\% confidence intervals from 10,000 case-level bootstrap resamples are reported in Appendix~\ref{app:statistical-results}.}
\label{tab:main-results}
\begin{tabular}{l cccc cc ccc cc}
\toprule
& \multicolumn{4}{c}{\textbf{Node identification}}
& \multicolumn{2}{c}{\textbf{Context}}
& \multicolumn{3}{c}{\textbf{Accuracy by query type}}
& \multicolumn{2}{c}{\textbf{Answer level}} \\
\cmidrule(lr){2-5}\cmidrule(lr){6-7}\cmidrule(lr){8-10}\cmidrule(lr){11-12}
\textbf{Method}
& DR $\uparrow$ & AR $\uparrow$ & UI $\downarrow$ & OI $\downarrow$
& SIC $\downarrow$ & CVER $\uparrow$
& Q1 $\uparrow$ & Q2 $\uparrow$ & Q3 $\uparrow$
& Final $\uparrow$ & Reuse $\downarrow$ \\
\midrule
Vector RAG      & --   & --   & --   & --   & 0.53 & 0.58 & 0.95 & 0.28 & 1.00 & 0.74 & 0.10 \\
Recency-RAG     & --   & --   & --   & --   & 0.47 & 0.66 & 1.00 & 0.39 & 0.95 & 0.78 & 0.04 \\
LLM-Only Update & 1.00 & --   & 0.68 & 0.00 & 0.49 & 0.64 & 1.00 & 0.47 & 0.42 & 0.63 & 0.03 \\
CUPMem-style    & 0.93 & --   & 1.00 & 0.00 & 0.40 & 0.76 & 0.97 & 0.43 & 1.00 & 0.80 & 0.03 \\
A-MEM-style     & --   & --   & --   & --   & 0.47 & 0.88 & 1.00 & 0.71 & 1.00 & 0.90 & 0.00 \\
\midrule
PACMI           & 1.00 & 1.00 & 0.00 & 0.00 & 0.34 & 0.83 & 1.00 & 0.97 & 1.00 & 0.99 & 0.00 \\
\bottomrule
\end{tabular}
\end{table*}

Table~\ref{tab:main-results} gives the comparison. Three observations matter, and only the first is a clean win.

First, at the answer level PACMI reaches 0.99 final accuracy against 0.90 for the strongest baseline, A-MEM-style, corresponding to 3 versus 29 errors over 300 paired queries. The difference is statistically significant under a two-sided exact McNemar test ($p=2.56\times10^{-6}$). The gap is concentrated entirely in Q2: both methods answer the current-state and historical queries at ceiling, so this benchmark's discriminative power lives primarily in stale-premise queries. Second, PACMI does \emph{not} dominate on retrieval: A-MEM-style attains higher current valid-evidence recall (0.88 versus 0.83), because link expansion pulls in more valid context than validity-aware ranking and because A-MEM-style has a larger maximum expanded context. PACMI is better on stale-in-context (0.34 versus 0.47), i.e.\ it admits less outdated material, but it pays for that in coverage. Third, stale reuse rounds to 0.00 for both PACMI and A-MEM-style, so on that metric we tie rather than lead.

The one clearly harmful design is destructive rewriting. LLM-Only Update matches PACMI on direct stale-node recall (1.00) but collapses on historical queries (0.42), because rewriting a memory in place destroys the record that a historical query needs. This reproduces, in a controlled setting, the degradation reported for continuously LLM-updated memory \cite{zhang2026faultymemories}, and it argues for non-destructive invalidation independently of any propagation mechanism.

\paragraph{Error analysis.}
PACMI's three errors occur after successful premise detection and primarily reflect answer-generation failures. In one case, the new-evidence node is absent from the retrieved top-five context, although the premise-correction block still supplies the current fact;
the remaining errors reflect insufficient grounding or incomplete execution of the expected answer behavior. LLM-Only Update's 58 historical-query errors primarily reflect loss of earlier values under in-place rewriting. Appendix~\ref{app:error-analysis} reports representative cases from both sets.

\subsection{Ablation}
\label{sec:ablation}

\begin{table}[t]
\centering
\footnotesize
\setlength{\tabcolsep}{2.5pt}
\caption{Ablation. AR, UI, and SIC are state- and context-level; Acc, Prem., and Hist.\ are answer-level. All variants have OI $=0.00$. Recall-at-depth and percentile 95\% case-bootstrap confidence intervals for all numeric cells are reported in Appendices~\ref{app:statistical-results} and~\ref{app:detailed-ablation}.}
\label{tab:ablation}
\begin{tabular}{@{}lcccccc@{}}
\toprule
\textbf{Variant} &
\textbf{AR $\uparrow$} & \textbf{UI $\downarrow$} & \textbf{SIC $\downarrow$} &
\textbf{Acc $\uparrow$} & \textbf{Prem. $\uparrow$} & \textbf{Hist. $\uparrow$} \\
\midrule
PACMI               & 1.00 & 0.00 & 0.34 & 0.99 & 1.00 & 1.00 \\
w/o Provenance      & 0.00 & 1.00 & 0.37 & 0.97 & 0.98 & 1.00 \\
w/o Cascade         & 0.00 & 1.00 & 0.41 & 0.96 & 0.98 & 0.99 \\
w/o Validity States & 1.00 & 0.00 & 0.56 & 0.97 & 0.98 & 1.00 \\
w/o Premise Check   & 1.00 & 0.00 & 0.34 & 0.83 & 0.80 & 1.00 \\
w/o Hist.\ Preserv. & 1.00 & 0.00 & 0.20 & 0.77 & 0.99 & 0.36 \\
\bottomrule
\end{tabular}
\end{table}

Table~\ref{tab:ablation} disables one component at a time. The largest answer-level drops occur when premise checking is removed (Final $0.99\!\to\!0.83$, Premise $1.00\!\to\!0.80$) or historical preservation is removed (Final $0.99\!\to\!0.77$, Hist.\ $1.00\!\to\!0.36$). On the 100 Q2 queries, PACMI makes 3 errors whereas the \emph{w/o Premise Checking} variant makes 50; the paired difference is significant under a two-sided exact McNemar test ($p=1.78\times10^{-13}$). This ablation removes both the conflict detector and the premise-aware answer prompt, so the result measures the combined premise-checking module rather than the detector alone.

Removing provenance use or cascading traversal drives affected recall to 0.00 and under-invalidates every case, while raising stale-in-context from 0.34 to 0.37 and 0.41, respectively. Their final-accuracy changes are much smaller: removing provenance produces 6 additional errors, and removing cascading produces 8 additional errors over 300 queries. For PACMI versus \emph{w/o Cascading Invalidation}, the error counts are 3 versus 11 and the two-sided exact McNemar test gives $p=0.0574$, which does not reach the 0.05 significance threshold. This non-significant result does not establish equivalence; it indicates that the current sample does not provide sufficient evidence for an answer-level cascading effect.

\paragraph{Intrinsic premise-detector performance.}
We evaluate the premise checker's structured \texttt{conflict} decision independently of final-answer generation, treating Q2 as the positive class and Q1/Q3 as negatives. Across the 300 queries, the checker produces 100 true positives, 200 true negatives, no false positives, and no false negatives, yielding precision, recall, and $F_1$ of 1.00. This result is conditional on the current controlled benchmark: the checker receives facts already separated by inferred graph states into outdated and current sets, Q2 directly assumes an outdated value, and Q3 explicitly asks about the past. It should therefore not be interpreted as evidence of perfect premise detection in unrestricted long-context settings.

\paragraph{Why cascading has limited answer-level effect.}
Each case contains six prior memories with a context budget of five, and PACMI retrieves the new-evidence node in 98.7\% of query contexts. The answer model therefore usually sees both the new evidence and the affected records even when downstream state labels are removed. Cascading propagation provides correct state assignment and auditability, but its paired final-accuracy difference is not significant at this scale ($p=0.0574$). Whether an answer-level gap emerges under larger memory stores or deeper dependency chains remains an open empirical question.

\subsection{Computational Cost}
\label{sec:cost}

\begin{table}[t]
\centering
\small
\setlength{\tabcolsep}{4pt}
\caption{Auxiliary LLM calls, excluding the one answer-generation call every method issues per query and the evaluation judge. ``Per update'' counts calls triggered by one new-evidence item; ``per query'' counts calls issued at question time. The last column totals auxiliary calls over the benchmark's 100 updates and 300 queries.}
\label{tab:cost}
\begin{tabular}{@{}lccc@{}}
\toprule
\textbf{Method} & \textbf{Per update} & \textbf{Per query} & \textbf{Total} \\
\midrule
Vector RAG      & 0 & 0 & 0 \\
Recency-RAG     & 0 & 0 & 0 \\
LLM-Only Update & 1 & 0 & 100 \\
CUPMem-style    & 1 & 0 & 100 \\
A-MEM-style     & 1 & 0 & 100 \\
\midrule
PACMI           & 1 & $\leq 1$ & 400 \\
\bottomrule
\end{tabular}
\end{table}

A mechanism that buys accuracy with inference is only interesting if the exchange rate is stated, so Table~\ref{tab:cost} reports it. Relation classification is a single batched call scoring all candidates at once, so PACMI matches the other update-aware methods at update time; the difference is the per-query premise check, which the baselines do not have. That check is skipped without an LLM call when no superseded record exists, but every case here contains one, so the bound is tight and PACMI issues four times the auxiliary calls of the next most expensive method. This is the correct denominator for our main result: 0.09 final accuracy over A-MEM-style for 300 extra calls, all at query time. Deployments that update often and query rarely therefore pay almost nothing, whereas query-dominated deployments should gate the premise check---for instance on whether the retrieved context contains any non-\textsc{active} record---rather than run it unconditionally. We measured calls, not wall-clock latency or tokens.

\section{Limitations}
\label{sec:limitations}

The dependency edges are supplied by annotation (A1). PACMI thus receives structural information that Vector RAG, Recency-RAG, and A-MEM-style do not, and inducing such edges from raw memory text is an unsolved problem in its own right. The comparison should be read as measuring what a system can do \emph{given} a provenance graph. The decisive missing experiment is a degradation study over graph quality: oracle edges, LLM-induced edges, and edges corrupted at controlled rates.

The benchmark uses a fixed two-hop topology with fixed edge semantics, so ground-truth labels are a deterministic function of topology (A3), over-invalidation cannot be exercised in the absence of downstream nodes that \emph{should not} be affected, and multi-path state resolution is never tested. Six prior memories per case also means retrieval is nearly saturated; the results say little about behavior at realistic memory scale.

Evaluation relies on an LLM judge drawn from the same model family as the answer generator, which risks self-preference bias, and each method was run once at temperature zero rather than over multiple seeds. We report the premise detector's precision, recall, and $F_1$ on the controlled query-type annotation as one judge-independent signal, but this evaluation is conditional on inferred graph states and templated query distinctions, and a human agreement study is still needed. Method context budgets also differ (A-MEM-style expands to eight entries against five elsewhere), which inflates its coverage-style metrics relative to the others. Finally, the stale-reuse judge scores misuse of directly stale memories and does not independently score every downstream validity state, and the cost accounting above counts LLM calls rather than measured latency or tokens, so it bounds rather than quantifies the overhead.

\section{Conclusion}
\label{sec:conclusion}

We formulated stale-memory resolution as dependency-aware validity management, gave a four-state validity lattice with typed-edge propagation, and released a diagnostic benchmark that separates memory-management behavior from final-answer behavior. On that benchmark PACMI attains the highest final accuracy, with a significant paired improvement over the strongest baseline. The graph-state-based premise checker achieves perfect precision, recall, and $F_1$ on the controlled query distribution, while non-destructive invalidation prevents the historical-query collapse observed under in-place LLM rewriting. Cascading propagation delivers correct downstream state assignment and auditability, but its paired final-answer difference is not statistically significant at the evaluated memory scale. Larger memory stores, deeper and branching dependency structures, graph-quality degradation studies, and human or cross-family evaluation are needed to determine whether these state-level benefits translate into broader answer-level gains.

\bibliography{references}

\clearpage
\appendix
\raggedbottom
\setcounter{secnumdepth}{1}

\section{Implementation Details}
\label{app:impl}

\paragraph{Models and API.}
We control the final-answer model, embedding model, and decoding temperature across methods. Method-specific context construction is retained; A-MEM-style may return up to eight entries, whereas the other methods return at most five. All methods use the same final-answer model and answer-level evaluation model: \texttt{deepseek-v4-pro}. Intermediate LLM operations use \texttt{deepseek-v4-flash}: memory rewriting in LLM-Only Update, slot-state consolidation in CUPMem-style, link evolution for newly inserted evidence in A-MEM-style, and relation classification and premise checking in PACMI and its ablations. Vector RAG and Recency-RAG have no intermediate LLM operation. All LLM calls use \texttt{temperature = 0}. Sentence embeddings are produced with \texttt{all-MiniLM-L6-v2} using SentenceTransformer.

The hosted endpoint is \url{https://api.deepseek.com}, accessed through the OpenAI-compatible Python client version \texttt{2.41.1}. The model names above are hosted service aliases rather than locally archived checkpoints, so exact output regeneration may be affected by server-side model updates. The released results and logs preserve the evaluated run.

\paragraph{Computing infrastructure.}
All answer-generation, relation-classification, premise-checking, and judging calls are issued to a hosted API, so no local accelerator is required; the reported results depend on the hosted API behavior associated with the model aliases listed above rather than on local hardware. Local computation is limited to sentence embedding and metric aggregation, and was run on \texttt{13th Gen Intel(R) Core(TM) i5-13600KF} with \texttt{31.85 GiB} of physical memory under \texttt{Microsoft Windows 11 Pro, version 10.0.26200, build 26200, 64-bit}. The software stack is Python \texttt{3.10.20} with \texttt{sentence-transformers} \texttt{5.5.1}, \texttt{torch} \texttt{2.12.0}, \texttt{numpy} \texttt{2.2.6}, and \texttt{scipy} \texttt{1.15.3}.

\paragraph{Randomness and seeds.}

All LLM calls use \texttt{temperature = 0}, and embedding computation is deterministic, so the inference pipeline is configured without stochastic decoding. The stochastic component of the statistical reanalysis is the bootstrap resampling used for confidence intervals (Section~\ref{app:stats}). We use a fixed seed of \texttt{20260728} and 10{,}000 case-level resamples. The seed and resampling routine are included in the released analysis script so that the reported intervals are reproducible.
\paragraph{Executable evaluation set.}
The executable experiments use a data file containing 100 memory-update cases and 300 queries. Each case contains six prior memories, one item of new evidence, and three queries covering the current state, a stale premise, and a historical state.

\paragraph{Case construction.}
Each benchmark case contains six prior memories, one newly observed evidence item, two prior-memory dependency edges, and exactly three test queries. The six memories are designed to include: (1) a directly affected memory, (2) at least one downstream memory derived from the affected information, (3) a related but still-valid memory, (4) a hard negative that is semantically similar but should not be updated, and (5) an unrelated distractor. The remaining entry provides additional supporting or contextual information.

We construct cases across five representative agent domains: Personal Profile, Location \& Preference, Work \& Project, Finance \& Recommendation, and Tool/Workflow State, with 20 cases in each domain.

\paragraph{Dependency and validity annotations.}
For the executable 100-case benchmark, the ground-truth state annotations use four labels:
\{
\textsc{active},
\textsc{superseded},
\textsc{needs\text{-}verification},
\textsc{historical\text{-}only}
\}.
For every case, $M_1$ is annotated as \textsc{superseded}, $M_2$ as \textsc{needs-verification}, $M_3$ as \textsc{historical-only}, and $M_4$--$M_6$ as \textsc{active}.

Each case contains two prior-memory dependency edges: $M_1 \xrightarrow{\textit{supports}} M_2$ and $M_2 \xrightarrow{\textit{used-by}} M_3$. Case-level ground truth provides the directly stale node IDs, downstream affected-node IDs, and expected post-update state assignments. The dataset does not provide a separate expected-repair-operation annotation. Historical information is preserved through timestamps together with the \textsc{historical-only} or \textsc{superseded} validity state.

As noted in assumption A3 (Section~\ref{sec:intro}), this construction makes the ground-truth state assignment a deterministic function of the case topology, and the topology is supplied to PACMI as input. We restate the consequence here so that it is visible wherever the annotations are read: affected-node recall, under-invalidation, and recall-at-depth measure whether propagation is implemented correctly, not whether provenance modeling confers an advantage over a method given the same graph. Section~\ref{app:future-benchmark} lists the benchmark extensions that would remove this property.

Each query is associated with an \texttt{expected\_behavior} annotation in the case-level \texttt{query\_ground\_truth} metadata, which is provided to the answer-level judge to assess whether the generated answer is correct. Current-valid-evidence and stale-context node sets are derived during evaluation from the case-level state annotations rather than stored as explicit per-query fields.

\paragraph{Query types.}
The benchmark contains three query types.
\emph{Current-state queries} (Q1) ask for the latest valid state of an entity, event, product, or system.
\emph{Stale-premise queries} (Q2) contain an assumption that was previously valid but has been contradicted or superseded by new evidence; a correct system should identify and correct the premise rather than answer under it.
\emph{Historical queries} (Q3) ask about information that was valid at an earlier time, and test whether a method preserves historical evidence without treating it as current.

We report performance separately for the three query types, because retrieving an outdated memory for a historical question is correct, whereas retrieving the same memory as current evidence constitutes stale-memory reuse.

\paragraph{Evaluation set and quality control.}
The executable evaluation uses all 100 cases as one fixed evaluation set. No development/test split or development-set threshold selection is used. Cases were reviewed for consistency among new evidence, dependency edges, state annotations, and expected query behavior. Each method is evaluated once on the fixed evaluation set using temperature-zero decoding; we report the metrics from this single run rather than a mean and standard deviation over multiple random seeds. See Section~\ref{app:stats} for the uncertainty quantification we report in place of seed variance.

\paragraph{Dataset statistics.}
Table~\ref{tab:dataset-stats} summarizes the benchmark statistics. Since every case contains six existing memories and one new evidence item, the memory and evidence columns report total counts rather than per-case averages.

\begin{table}[t]
\centering
\small
\caption{Statistics of the stale-memory dependency benchmark.}
\label{tab:dataset-stats}
\resizebox{\columnwidth}{!}{
\begin{tabular}{lrrrrr}
\toprule
\textbf{Domain} &
\textbf{Cases} &
\textbf{Old Mem.} &
\textbf{New Evid.} &
\textbf{Queries} &
\textbf{Affected} \\
\midrule
Personal Profile          & 20  & 120 & 20  & 60  & 40  \\
Location \& Preference     & 20  & 120 & 20  & 60  & 40  \\
Work \& Project            & 20  & 120 & 20  & 60  & 40  \\
Finance \& Recommendation  & 20  & 120 & 20  & 60  & 40  \\
Tool / Workflow State      & 20  & 120 & 20  & 60  & 40  \\
\midrule
Total                     & 100 & 600 & 100 & 300 & 200 \\
\bottomrule
\end{tabular}}
\end{table}

\paragraph{Candidate selection.}
Each benchmark case contains six prior memories, and PACMI uses a candidate budget of $k=6$. Candidate scoring therefore determines the ordering of prior memories but does not prune any prior-memory node in the current benchmark.
When new evidence $e_{t+1}$ arrives, PACMI first selects a set of candidate prior memories that may be directly related to the update. For each prior memory $m_i$, candidate selection combines embedding-based cosine similarity, token-level Jaccard overlap, and a bonus for matching non-\textit{other} semantic slots:
\[
\begin{aligned}
s_{\mathrm{cand}}(e_{t+1},m_i)
={}&0.82\,s_{\mathrm{cos}}(e_{t+1},m_i)\\
&+0.18\,s_{\mathrm{tok}}(e_{t+1},m_i) \\
&+0.08\,\mathbb{I}[\sigma_e=\sigma_i\neq\textit{other}],
\end{aligned}
\]
where $\sigma_e=\operatorname{slot}(e_{t+1})$, $\sigma_i=\operatorname{slot}(m_i)$, and $s_{\mathrm{tok}}$ denotes token-level Jaccard overlap. The implementation retains up to six highest-ranked prior memories as relation-classification candidates. Candidate selection is designed for recall and does not itself determine whether a memory is stale.

\paragraph{Key hyperparameters.}
Table~\ref{tab:hyperparams} reports the main shared and method-specific hyperparameters. PACMI, all PACMI ablations, Vector RAG, Recency-RAG, LLM-Only Update, and CUPMem-style retrieve at most $k{=}5$ entries for response generation. A-MEM-style first selects up to five seed notes and then expands their one-hop linked neighbors, with a maximum final context size of eight. This budget difference is a known confound for the coverage-style context metrics and is discussed in Section~\ref{sec:limitations}.

\begin{table}[t]
\centering
\small
\caption{Key hyperparameters used in the main experiments.}
\label{tab:hyperparams}
\resizebox{\columnwidth}{!}{
\begin{tabular}{llc}
\toprule
\textbf{Method} & \textbf{Parameter} & \textbf{Value} \\
\midrule
PACMI, Vector RAG, Recency-RAG,
    & top-$k$ (final context) & 5 \\
LLM-Only Update, CUPMem-style
    &  &  \\
A-MEM-style
    & seed top-$k$ & 5 \\
A-MEM-style
    & maximum expanded context & 8 \\
Recency-RAG
    & similarity weight & 0.7 \\
Recency-RAG
    & recency weight & 0.3 \\
All LLM calls
    & temperature & 0 \\
\midrule
All methods
    & final-answer model & deepseek-v4-pro \\
All except Vector RAG,
    & auxiliary model & deepseek-v4-flash \\
Recency-RAG
    &  &  \\
All methods
    & answer-level judge model & deepseek-v4-pro \\
All methods
    & embedding model & all-MiniLM-L6-v2 \\
\bottomrule
\end{tabular}}
\end{table}

\paragraph{Retrieval scoring.}
Vector RAG ranks entries by cosine similarity. Recency-RAG uses
$0.7s_{\mathrm{sim}}+0.3s_{\mathrm{rec}}$. LLM-Only Update first rewrites the
memory store and then applies cosine-similarity retrieval. CUPMem-style first applies query-dependent eligibility filtering: current-oriented queries exclude superseded records, whereas historical queries additionally admit superseded previous values. Eligible records are then ranked using cosine similarity together with memory-state, slot-state, new-evidence, and recency biases. A-MEM-style augments cosine similarity with slot, link, new-evidence, and recency biases before expanding one-hop linked notes.

For PACMI, the query-time score is cosine similarity plus a validity-state bias, a $0.06$ new-evidence bonus, a $0.05$ matching-slot bonus, and a $0.04$ normalized-recency bonus. For current-state and stale-premise queries, the state biases for \textsc{active}, \textsc{needs-verification}, \textsc{historical-only}, and \textsc{superseded} are $0.12$, $0.04$, $-0.10$, and $-0.22$, respectively; this ordering is the operational definition of the validity lattice given in Section~\ref{sec:problem}. For historical queries, the corresponding biases are $0.04$, $0.04$, $0.12$, and $0.12$.

\paragraph{Baseline capability boundaries.}
Vector RAG and Recency-RAG perform retrieval without an explicit invalidation decision. LLM-Only Update makes local UPDATE/CONTRADICT decisions but does not propagate downstream effects. CUPMem-style makes slot-level supersession decisions and preserves previous values for historical queries, but it does not traverse dependency edges. A-MEM-style constructs and expands semantic links without explicit validity propagation. PACMI combines direct relation decisions with dependency propagation, state-aware retrieval, and premise checking.

\section{Formal Metric Definitions}
\label{app:metrics}

\paragraph{Stale- and affected-node identification.}
For case $c$, let $\mathcal{D}_c$ and $\mathcal{A}_c$ be the ground-truth direct and affected sets and hats denote predictions. We report case-macro recall:
\begin{equation}
\begin{aligned}
\mathrm{DR}&=\frac{1}{|\mathcal C_D|}\sum_{c\in\mathcal C_D}
\frac{|\mathcal D_c\cap\widehat{\mathcal D}_c|}{|\mathcal D_c|},\\
\mathrm{AR}&=\frac{1}{|\mathcal C_A|}\sum_{c\in\mathcal C_A}
\frac{|\mathcal A_c\cap\widehat{\mathcal A}_c|}{|\mathcal A_c|}.
\end{aligned}
\end{equation}

For methods with explicit non-current decisions, let $\mathcal I_c=\mathcal D_c\cup\mathcal A_c$. Predictions come from UPDATE/CONTRADICT decisions for LLM-Only Update, slot supersession for CUPMem-style, and direct plus propagated predictions for PACMI:
\begin{equation}
\begin{aligned}
\widehat{\mathcal O}_c&=\widehat{\mathcal I}_c\setminus\mathcal I_c,&
\mathrm{OI}&=\frac{1}{|\mathcal C_I|}\sum_{c\in\mathcal C_I}\mathbb I[\widehat{\mathcal O}_c\ne\emptyset],\\
\widehat{\mathcal U}_c&=\mathcal I_c\setminus\widehat{\mathcal I}_c,&
\mathrm{UI}&=\frac{1}{|\mathcal C_I|}\sum_{c\in\mathcal C_I}\mathbb I[\widehat{\mathcal U}_c\ne\emptyset].
\end{aligned}
\end{equation}
AR is not applicable without explicit downstream prediction; OI and UI are not applicable without an invalidation or supersession decision.

\paragraph{Context-level metrics.}
For query $q$, let $\mathcal R_q$ be retrieved IDs, $\mathcal S_q$ direct or affected ground-truth IDs, and $\mathcal G_q$ the active prior memories plus new evidence. Over current-oriented queries $\mathcal Q_{\mathrm{cur}}$ (Q1/Q2),
\begin{equation}
\begin{aligned}
\mathrm{SIC}&=\frac{1}{|\mathcal Q_{\mathrm{cur}}|}
\sum_{q\in\mathcal Q_{\mathrm{cur}}}
\frac{|\mathcal R_q\cap\mathcal S_q|}{|\mathcal R_q|},\\
\mathrm{CVER}&=\frac{1}{|\mathcal Q_{\mathrm{cur}}|}
\sum_{q\in\mathcal Q_{\mathrm{cur}}}
\frac{|\mathcal R_q\cap\mathcal G_q|}{|\mathcal G_q|}.
\end{aligned}
\end{equation}
Q3 is excluded because superseded or historical-only records may be appropriate historical evidence. Note that $|\mathcal R_q|$ appears in the SIC denominator and $\mathcal R_q$ in the CVER numerator, so both metrics are sensitive to the context budget; methods should be compared at a matched budget.

\paragraph{Answer-level metrics.}
\textbf{Final Accuracy} follows the annotated current or historical behavior. \textbf{Stale Reuse} records whether an answer treats a directly stale prior memory as current; it does not independently score every downstream state. \textbf{Premise Resistance} is accuracy in correcting the stale assumption in Q2, and \textbf{Historical Preservation} is Q3 final accuracy. The fixed \texttt{deepseek-v4-pro} judge applies the same temperature-zero rubric to all methods.

\paragraph{Recall at dependency depth.}
Dividing affected memories by their shortest dependency distance from the directly conflicted node, depth-1 nodes directly depend on the directly stale memory and depth-2 nodes are reached through one additional dependency edge. The 100-case benchmark contains 100 affected nodes at each depth. We report
\[
\mathrm{Recall@Depth}(d)
=
\frac{
|\mathcal{A}_d\cap\widehat{\mathcal{A}}_d|
}{
|\mathcal{A}_d|
}.
\]

\section{Evaluation Protocol}
\label{app:evaluation-protocol}

\paragraph{State- and context-level construction.}
For each case, the ground-truth non-current set is the union of direct stale and downstream affected nodes. LLM-Only Update predicts all prior memories classified as UPDATE or CONTRADICT; CUPMem-style predicts the union of its slot-level superseded IDs; PACMI predicts its direct stale and propagated affected IDs. Over-Invalidation records whether the predicted set contains any node outside ground truth, and Under-Invalidation records whether any ground-truth non-current node is missed. Vector RAG, Recency-RAG, and A-MEM-style make no explicit invalidation decision, so these metrics are not applicable. Affected Recall is applicable only to PACMI and propagation-enabled ablations; an empty placeholder field for a non-propagating method is not interpreted as a prediction.

Stale-in-Context and Current Valid-Evidence Recall are averaged over Q1/Q2 only. For these queries, stale context consists of directly stale or downstream affected ground-truth IDs, while current valid evidence consists of active prior-memory IDs plus the new-evidence ID. Q3 receives \texttt{NA} because superseded and historical-only nodes may be appropriate evidence for an explicitly historical query. Method-inapplicable values are stored as \texttt{NA} and displayed as ``--''; a numeric zero is used only when a metric is applicable and the measured error set is empty.
New-Evidence Coverage is the mean over all 300 query rows of $\mathbb{I}[\text{new-evidence ID is included in the retrieved context}]$.

\paragraph{Answer-level evaluation.}
The answer-level judge uses \texttt{deepseek-v4-pro} with temperature zero. For each query, it receives the query, final answer, annotated expected behavior, retrieved memory snippets, stale-memory annotations, and new evidence. The complete system and user prompts are in Section~\ref{app:prompts}. Because the judge is drawn from the same model family as the answer generator, we additionally report a judge-independent measurement of the premise detector in Section~\ref{app:premise-detector}.

\section{Intrinsic Evaluation of the Premise Detector}
\label{app:premise-detector}

The premise checker emits one structured \texttt{conflict} decision for each query. We evaluate this decision independently of final-answer generation, using the benchmark query-type annotation as ground truth: Q2 is the positive class because it contains a stale premise, while Q1 and Q3 are negative because they ask about the current state and historical state, respectively.

\begin{table}[t]
\centering
\small
\caption{Intrinsic premise-detector results. Q2 is the positive class; Q1 and Q3 are negative.}
\label{tab:supp-premise-detector}
\begin{tabular}{rrrrrrr}
\toprule
\textbf{TP} &
\textbf{FP} &
\textbf{FN} &
\textbf{TN} &
\textbf{Prec.} &
\textbf{Rec.} &
\textbf{$F_1$} \\
\midrule
100 & 0 & 0 & 200 & 1.000 & 1.000 & 1.000 \\
\bottomrule
\end{tabular}
\end{table}

All 100 Q2 queries are classified as conflicts, while all 100 Q1 and 100 Q3 queries are classified as non-conflicts. The detector therefore has no false positives and no false negatives on this benchmark. This result is conditional on the evaluated pipeline: the checker receives facts already separated by the inferred graph states into outdated and current sets, Q2 directly assumes an outdated value, and Q3 explicitly asks about the past. The result demonstrates correct premise-conflict classification on the current controlled distribution, not perfect detection in unrestricted or naturalistic long-context settings.

\section{Statistical Protocol}
\label{app:stats}

Each method is run once at temperature zero, so the reported uncertainty is over benchmark cases rather than over inference seeds. For every numeric cell reported in Tables~\ref{tab:main-results} and~\ref{tab:ablation}, we compute a percentile 95\% confidence interval from 10{,}000 case-level bootstrap resamples using seed \texttt{20260728}. Each bootstrap replicate samples 100 cases with replacement, and all three queries belonging to a sampled case move together. Structural quantities retain their original case-level definitions within each replicate. Method-inapplicable cells marked ``--'' are not bootstrapped.

For the three reported paired comparisons, we use a two-sided exact McNemar test on the discordant paired outcomes. The comparisons are PACMI versus A-MEM-style on final accuracy over all 300 queries, PACMI versus \emph{w/o Cascading Invalidation} on final accuracy over all 300 queries, and PACMI versus \emph{w/o Premise Checking} on Q2 accuracy over the 100 stale-premise queries. We use the conventional $\alpha=0.05$ threshold and report the full discordant counts rather than only the significance decision. A non-significant result is not interpreted as evidence of equivalence.

The three exact $p$-values are reported without multiplicity adjustment. They should therefore be interpreted as per-comparison evidence rather than as family-wise error-controlled decisions.

\section{Statistical Reanalysis Results}
\label{app:statistical-results}

All values below are reported as estimate [percentile 95\% confidence interval]. Intervals use 10{,}000 case-level bootstrap resamples with seed \texttt{20260728}. Only numeric cells present in Tables~\ref{tab:main-results} and~\ref{tab:ablation} are included.

\begin{table*}[t]
\centering
\scriptsize
\setlength{\tabcolsep}{3pt}
\caption{Case-bootstrap intervals for the node- and context-level cells of the main-results table. ``--'' denotes a method-inapplicable metric.}
\label{tab:supp-main-node-context-ci}
\resizebox{\textwidth}{!}{
\begin{tabular}{lcccccc}
\toprule
\textbf{Method} &
\textbf{DR} &
\textbf{AR} &
\textbf{UI} &
\textbf{OI} &
\textbf{SIC} &
\textbf{CVER} \\
\midrule
Vector RAG
& -- & -- & -- & --
& 0.534 [0.520, 0.548]
& 0.583 [0.565, 0.600] \\
Recency-RAG
& -- & -- & -- & --
& 0.473 [0.456, 0.489]
& 0.659 [0.639, 0.680] \\
LLM-Only Update
& 1.000 [1.000, 1.000]
& --
& 0.680 [0.590, 0.770]
& 0.000 [0.000, 0.000]
& 0.487 [0.468, 0.505]
& 0.641 [0.619, 0.665] \\
CUPMem-style
& 0.930 [0.870, 0.980]
& --
& 1.000 [1.000, 1.000]
& 0.000 [0.000, 0.000]
& 0.396 [0.385, 0.407]
& 0.755 [0.741, 0.769] \\
A-MEM-style
& -- & -- & -- & --
& 0.467 [0.455, 0.479]
& 0.882 [0.848, 0.916] \\
PACMI
& 1.000 [1.000, 1.000]
& 1.000 [1.000, 1.000]
& 0.000 [0.000, 0.000]
& 0.000 [0.000, 0.000]
& 0.338 [0.320, 0.355]
& 0.828 [0.806, 0.850] \\
\bottomrule
\end{tabular}}
\end{table*}

\begin{table*}[t]
\centering
\scriptsize
\setlength{\tabcolsep}{3pt}
\caption{Case-bootstrap intervals for the query-type and answer-level cells of the main-results table.}
\label{tab:supp-main-answer-ci}
\resizebox{\textwidth}{!}{
\begin{tabular}{lccccc}
\toprule
\textbf{Method} &
\textbf{Q1 Accuracy} &
\textbf{Q2 Accuracy} &
\textbf{Q3 Accuracy} &
\textbf{Final Accuracy} &
\textbf{Stale Reuse} \\
\midrule
Vector RAG
& 0.950 [0.900, 0.990]
& 0.280 [0.190, 0.370]
& 1.000 [1.000, 1.000]
& 0.743 [0.710, 0.777]
& 0.100 [0.070, 0.133] \\
Recency-RAG
& 1.000 [1.000, 1.000]
& 0.390 [0.290, 0.490]
& 0.950 [0.900, 0.990]
& 0.780 [0.743, 0.817]
& 0.040 [0.020, 0.063] \\
LLM-Only Update
& 1.000 [1.000, 1.000]
& 0.470 [0.370, 0.570]
& 0.420 [0.320, 0.520]
& 0.630 [0.580, 0.680]
& 0.027 [0.010, 0.047] \\
CUPMem-style
& 0.970 [0.930, 1.000]
& 0.430 [0.330, 0.530]
& 1.000 [1.000, 1.000]
& 0.800 [0.767, 0.837]
& 0.033 [0.013, 0.057] \\
A-MEM-style
& 1.000 [1.000, 1.000]
& 0.710 [0.620, 0.800]
& 1.000 [1.000, 1.000]
& 0.903 [0.873, 0.933]
& 0.003 [0.000, 0.010] \\
PACMI
& 1.000 [1.000, 1.000]
& 0.970 [0.930, 1.000]
& 1.000 [1.000, 1.000]
& 0.990 [0.977, 1.000]
& 0.000 [0.000, 0.000] \\
\bottomrule
\end{tabular}}
\end{table*}

\begin{table*}[t]
\centering
\scriptsize
\setlength{\tabcolsep}{3pt}
\caption{Case-bootstrap intervals for all numeric cells of the ablation table.}
\label{tab:supp-ablation-ci}
\resizebox{\textwidth}{!}{
\begin{tabular}{lcccccc}
\toprule
\textbf{Variant} &
\textbf{Affected Recall} &
\textbf{Under-Invalidation} &
\textbf{Stale-in-Context} &
\textbf{Final Accuracy} &
\textbf{Premise Resistance} &
\textbf{Historical Preservation} \\
\midrule
PACMI
& 1.000 [1.000, 1.000]
& 0.000 [0.000, 0.000]
& 0.338 [0.320, 0.355]
& 0.990 [0.977, 1.000]
& 1.000 [1.000, 1.000]
& 1.000 [1.000, 1.000] \\
w/o Provenance Graph
& 0.000 [0.000, 0.000]
& 1.000 [1.000, 1.000]
& 0.369 [0.351, 0.387]
& 0.970 [0.950, 0.987]
& 0.980 [0.950, 1.000]
& 1.000 [1.000, 1.000] \\
w/o Cascading Invalidation
& 0.000 [0.000, 0.000]
& 1.000 [1.000, 1.000]
& 0.413 [0.399, 0.426]
& 0.963 [0.940, 0.983]
& 0.980 [0.950, 1.000]
& 0.990 [0.970, 1.000] \\
w/o Validity States
& 1.000 [1.000, 1.000]
& 0.000 [0.000, 0.000]
& 0.561 [0.549, 0.572]
& 0.970 [0.947, 0.990]
& 0.980 [0.950, 1.000]
& 1.000 [1.000, 1.000] \\
w/o Premise Checking
& 1.000 [1.000, 1.000]
& 0.000 [0.000, 0.000]
& 0.338 [0.320, 0.355]
& 0.833 [0.800, 0.867]
& 0.800 [0.720, 0.870]
& 1.000 [1.000, 1.000] \\
w/o Historical Preservation
& 1.000 [1.000, 1.000]
& 0.000 [0.000, 0.000]
& 0.200 [0.200, 0.200]
& 0.767 [0.733, 0.800]
& 0.990 [0.970, 1.000]
& 0.360 [0.270, 0.460] \\
\bottomrule
\end{tabular}}
\end{table*}

\begin{table*}[t]
\centering
\small
\caption{Two-sided exact McNemar tests for the three reported paired comparisons. Discordant counts are reported as PACMI-only correct / comparison-method-only correct.}
\label{tab:supp-mcnemar}
\resizebox{\textwidth}{!}{
\begin{tabular}{llcccc}
\toprule
\textbf{Comparison} &
\textbf{Scope} &
\textbf{Errors} &
\textbf{Discordant} &
\textbf{Exact $p$} &
\textbf{$p<0.05$} \\
\midrule
PACMI vs A-MEM-style
& Q1+Q2+Q3
& 3 / 29
& 29 / 3
& $2.55601\times10^{-6}$
& yes \\
PACMI vs w/o Cascading Invalidation
& Q1+Q2+Q3
& 3 / 11
& 11 / 3
& 0.057373
& no \\
PACMI vs w/o Premise Checking
& Q2
& 3 / 50
& 48 / 1
& $1.77636\times10^{-13}$
& yes \\
\bottomrule
\end{tabular}}
\end{table*}

PACMI significantly outperforms A-MEM-style on final accuracy and the \emph{w/o Premise Checking} variant on Q2 accuracy. The final-accuracy comparison with \emph{w/o Cascading Invalidation} does not reach the 0.05 threshold. This non-significant result does not establish equivalence.

\paragraph{Per-method New-Evidence Coverage.}
New-Evidence Coverage is the fraction of all 300 query contexts that contain the new-evidence node $N_1$. Table~\ref{tab:supp-new-evidence-coverage} reports the value separately for every main method.

\begin{table}[t]
\centering
\small
\caption{New-Evidence Coverage by main method.}
\label{tab:supp-new-evidence-coverage}
\begin{tabular}{lc}
\toprule
\textbf{Method} & \textbf{New-Evidence Coverage} \\
\midrule
Vector RAG       & 0.843 \\
Recency-RAG      & 1.000 \\
LLM-Only Update  & 0.863 \\
CUPMem-style     & 1.000 \\
A-MEM-style      & 1.000 \\
PACMI            & 0.987 \\
\bottomrule
\end{tabular}
\end{table}

\section{Ablation Implementations}
\label{app:ablations}

Each ablation disables one PACMI component while leaving the remaining pipeline unchanged.

\begin{table}[t]
\centering
\small
\caption{Implementation of the PACMI ablations.}
\label{tab:ablation-impl}
\resizebox{\columnwidth}{!}{
\begin{tabular}{lp{0.68\columnwidth}}
\toprule
\textbf{Ablation} & \textbf{Implementation} \\
\midrule
w/o provenance graph
    & Disables the use of incoming dependency edges for distinguishing directly
      replaced memories from downstream dependency records and disables all graph
      propagation. Relation classification, deterministic consistency constraints,
      direct stale detection, and direct-state assignment remain enabled. \\
w/o cascading invalidation
    & Retains the graph and direct stale-memory detection but disables
      traversal from directly stale nodes to downstream dependent nodes. \\
w/o validity states
    & Removes validity-state filtering, state-dependent ranking biases, and
      validity-state labels from the retrieved context. Other query-time
      ranking signals remain unchanged. \\
w/o premise checking
    & Skips premise detection and always uses the neutral answer-generation
      prompt. \\
w/o historical preservation
    & Excludes superseded and historical-only prior memories from retrieval,
      including for historical queries. \\
\bottomrule
\end{tabular}}
\end{table}

\section{Detailed Ablation Results}
\label{app:detailed-ablation}

\begin{table}[t]
\centering
\small
\caption{State- and context-level ablation results for PACMI. Higher is better for Affected Recall and New-Evidence Coverage; lower is better for Over-Invalidation, Under-Invalidation, and Stale-in-Context. Per assumption A3, the Affected Recall and Under-Invalidation columns verify propagation correctness rather than comparative advantage.}
\label{tab:supp-ablation-state}
\resizebox{\columnwidth}{!}{
\begin{tabular}{lccccc}
\toprule
\textbf{Variant} &
\textbf{Aff. Rec. $\uparrow$} &
\textbf{Over-Inv. $\downarrow$} &
\textbf{Under-Inv. $\downarrow$} &
\textbf{SIC $\downarrow$} &
\textbf{New-Ev. Cov. $\uparrow$} \\
\midrule
PACMI                        & 1.00 & 0.00 & 0.00 & 0.34 & 0.99 \\
w/o Provenance Graph         & 0.00 & 0.00 & 1.00 & 0.37 & 0.99 \\
w/o Cascading Invalidation   & 0.00 & 0.00 & 1.00 & 0.41 & 0.99 \\
w/o Validity States          & 1.00 & 0.00 & 0.00 & 0.56 & 0.96 \\
w/o Premise Checking         & 1.00 & 0.00 & 0.00 & 0.34 & 0.99 \\
w/o Historical Preservation  & 1.00 & 0.00 & 0.00 & 0.20 & 1.00 \\
\bottomrule
\end{tabular}}
\end{table}

\paragraph{Over- versus under-invalidation.}
Among methods that make an explicit invalidation decision, none over-invalidates an active memory in the current benchmark. This is a property of the benchmark rather than of the methods: because every downstream node of a directly replaced memory is annotated as non-current, the benchmark contains no node that is dependency-connected to a stale record yet should remain \textsc{active}, so the over-invalidation metric cannot be exercised. The observable difference therefore lies entirely in under-invalidation: LLM-Only Update misses at least one non-current node in 68\% of cases, and CUPMem-style in all cases, because neither propagates downstream effects.

\begin{table}[t]
\centering
\small
\caption{Answer-level ablation results for PACMI. Higher is better except for stale memory reuse.}
\label{tab:supp-ablation-answer}
\resizebox{\columnwidth}{!}{
\begin{tabular}{lccc}
\toprule
\textbf{Variant} &
\textbf{Final Acc. $\uparrow$} &
\textbf{Stale Reuse $\downarrow$} &
\textbf{Premise Resist. $\uparrow$} \\
\midrule
PACMI                        & 0.99 & 0.00 & 1.00 \\
w/o Provenance Graph         & 0.97 & 0.01 & 0.98 \\
w/o Cascading Invalidation   & 0.96 & 0.01 & 0.98 \\
w/o Validity States          & 0.97 & 0.02 & 0.98 \\
w/o Premise Checking         & 0.83 & 0.02 & 0.80 \\
w/o Historical Preservation  & 0.77 & 0.00 & 0.99 \\
\bottomrule
\end{tabular}}
\end{table}

\begin{table}[t]
\centering
\small
\caption{Ablation results for PACMI: final-answer accuracy by query type.}
\label{tab:supp-ablation-query}
\resizebox{\columnwidth}{!}{
\begin{tabular}{lccc}
\toprule
\textbf{Variant} &
\textbf{Current} &
\textbf{Stale Premise} &
\textbf{Historical} \\
\midrule
PACMI                        & 1.00 & 0.97 & 1.00 \\
w/o Provenance Graph         & 1.00 & 0.91 & 1.00 \\
w/o Cascading Invalidation   & 0.99 & 0.91 & 0.99 \\
w/o Validity States          & 0.97 & 0.94 & 1.00 \\
w/o Premise Checking         & 1.00 & 0.50 & 1.00 \\
w/o Historical Preservation  & 1.00 & 0.94 & 0.36 \\
\bottomrule
\end{tabular}}
\end{table}

\begin{table}[t]
\centering
\small
\caption{Affected-node recall at each dependency depth. Variants without graph propagation miss downstream affected nodes at both depths. As with Affected Recall, these figures verify propagation correctness under assumption A3.}
\label{tab:supp-depth}
\resizebox{\columnwidth}{!}{
\begin{tabular}{lcc}
\toprule
\textbf{Variant} &
\textbf{Recall@Depth-1} $\uparrow$ &
\textbf{Recall@Depth-2} $\uparrow$ \\
\midrule
PACMI                        & 1.00 & 1.00 \\
w/o Provenance Graph         & 0.00 & 0.00 \\
w/o Cascading Invalidation   & 0.00 & 0.00 \\
w/o Validity States          & 1.00 & 1.00 \\
w/o Premise Checking         & 1.00 & 1.00 \\
w/o Historical Preservation  & 1.00 & 1.00 \\
\bottomrule
\end{tabular}}
\end{table}

\paragraph{Analysis by query type.}
Current-state queries measure whether stale information is appropriately handled when producing the current answer; stale-premise queries measure whether the system identifies and corrects an invalid assumption; historical queries measure whether past information is preserved and retrieved under the correct temporal context. PACMI obtains 1.00, 0.97, and 1.00 respectively. Its observed advantage in the reported comparison is concentrated on stale-premise queries. CUPMem-style preserves historical answers but performs less reliably on stale-premise queries under the main evaluation setting. Within PACMI, the \emph{w/o Premise Checking} variant retains the same retrieval and graph states but removes both premise-conflict detection and the premise-aware correction block: Q2 accuracy falls from 0.97 to 0.50, while Q1 and Q3 are unchanged. This ablation demonstrates the importance of the combined premise-checking module but does not isolate detection from answer-side prompting.

\section{Representative Error Analysis}
\label{app:error-analysis}

The analysis script enumerated all three PACMI final-answer errors and all 58 LLM-Only Update errors on Q3. We selected two representative examples from each set to illustrate the observed failure patterns across cases and domains. The quantitative error counts are computed from the logs; the mechanism descriptions below are qualitative interpretations of the corresponding retrieved contexts, rewritten memories, final answers, and judge reasons.

\paragraph{PACMI: \texttt{case\_0022/Q2}.}
The query assumes that the user's current home is London, whereas the
new evidence states that the current city is San Francisco. PACMI's
premise checker correctly identifies London as outdated and returns
San Francisco as the current fact. Although $N_1$ is absent from the
top-five retrieved context, the premise-correction block still supplies
the current fact to the answer generator. The generated answer rejects
London as current but fails to state San Francisco. This is therefore
best interpreted as a generation or instruction-following failure under
reduced retrieval support, rather than as a pure premise-detection failure
or a pure retrieval-coverage failure.

\paragraph{PACMI: \texttt{case\_0072/Q2}.}
The query assumes a tax-sensitive-income investment profile, whereas the current profile is diversified growth. PACMI retrieves $N_1$ and correctly identifies and repairs the premise, but the generator proceeds to give a generic diversified-growth recommendation. The expected behavior requires acknowledging that the memory contains no specific allocation or fund recommendation for the new profile. This is a generation-grounding failure rather than a detection or retrieval failure.

\paragraph{LLM-Only Update: \texttt{case\_0001/Q3}.}
The historical query asks which profession preceded data scientist. LLM-Only Update rewrites the original profession memory from accountant to data scientist. The previous value is therefore no longer available, and the final answer incorrectly reports that the earlier profession is unspecified. This illustrates destructive loss of historical evidence under in-place rewriting.

\paragraph{LLM-Only Update: \texttt{case\_0021/Q3}.}
The historical query asks which city preceded Singapore. LLM-Only Update rewrites the original home city from Berlin to Singapore, after which the system can no longer recover Berlin as the historical answer. This is the same destructive-rewriting mechanism in a different domain.

\section{Benchmark Extensions That Remove the Current Assumptions}
\label{app:future-benchmark}

We list these explicitly so that the released benchmark can be extended along known axes.
\textbf{(i) Topology.} Varying dependency depth beyond two, adding branching and confluence, and adding downstream nodes that are dependency-connected but semantically unaffected, would decouple ground-truth labels from topology and make over-invalidation and multi-path state resolution measurable.
\textbf{(ii) Graph quality.} Comparing oracle edges, edges induced from memory text, and edges corrupted at controlled rates would quantify how much of PACMI's behavior depends on assumption A1.
\textbf{(iii) Memory scale.} Increasing the number of prior memories per case well beyond the context budget would test the prediction that propagation's answer-level value grows with retrieval pressure.
\textbf{(iv) Evaluation.} Multiple answer-generation backbones, a cross-family judge, and a human agreement study would address the judge-dependence noted in Section~\ref{sec:limitations}.

\clearpage
\section{Prompt Templates}
\label{app:prompts}

PACMI contains three LLM-driven system components: relation classification, premise checking, and answer generation. The evaluation additionally uses an LLM-based answer-level judge. Relation classification, premise checking, and answer judging require structured JSON output, whereas answer generation returns plain text. All calls use \texttt{temperature = 0} and a task-specific system message. Angle-bracketed fields are populated at runtime. Line wrapping, example items, and dynamically rendered lists are normalized below without changing the instructional content.

\paragraph{Relation Classification Prompt.}
The relation classifier uses \texttt{deepseek-v4-flash}. Its system message is:
\textit{``You are a strict memory-relation classifier. Return JSON only.''}

\begin{lstlisting}[basicstyle=\footnotesize\ttfamily,breaklines=true,
  frame=single,xleftmargin=0pt]
You are a memory relation classifier.
Task: classify the relation between the new evidence and each old memory. Return JSON only.

Allowed labels: DUPLICATE, REFINE, UPDATE, CONTRADICT, TEMPORALLY_COMPATIBLE, UNRELATED.

Definitions:
- UPDATE: the new evidence directly replaces an old current fact in the same slot.
- CONTRADICT: the new evidence directly conflicts with an old current fact in the same slot.
- REFINE: the new evidence adds detail without invalidating the old memory.
- TEMPORALLY_COMPATIBLE: both memories can be true at different times.
- DUPLICATE: both memories express nearly the same fact.
- UNRELATED: no substantive relation.

Constraints:
- Use UPDATE or CONTRADICT only when the old memory itself is the current fact replaced by the new evidence.
- Treat markers such as current, active, on file, latest saved, still lists, still says, and still treats as assertions of a current fact. If new evidence explicitly says now/current/changed/updated and gives a different value in the same single-value slot, classify that old current fact as UPDATE or CONTRADICT, not TEMPORALLY_COMPATIBLE or REFINE.
- Use TEMPORALLY_COMPATIBLE only when the old memory is explicitly historical, using wording such as previous, earlier, former, used to, was previously, or historical. An older timestamp by itself does not make a current-fact statement historical.
- Stable user preferences, output-format preferences, general rules, reusable checklists, templates, and advice style should not become stale merely because a profession, location, project, or tool changed.
- If an old memory is a plan, draft, recommendation, or historical record based on an old fact, prefer TEMPORALLY_COMPATIBLE or REFINE; downstream impact is handled separately by graph propagation.
- If the old memory and new evidence are not in the same slot, use UNRELATED.

New evidence:
id=<id> | time=<time> | slot=<slot> | text=<text>

Candidate old memories:
<numbered list of id, time, slot, and text>

Return JSON in this schema:
{
  "relations": [
    {"memory_id": "M1", "relation": "UPDATE",
     "reason": "..."}
  ]
}
\end{lstlisting}

The returned labels are subject to deterministic consistency checks. In particular, \textsc{update} and \textsc{contradict} are accepted only for matching single-valued slots, and stable contextual details are not marked directly stale. Graph propagation handles downstream plans, claims, and decisions separately from direct relation classification.

\paragraph{Premise Checking Prompt.}
Premise checking runs after graph-state propagation and uses \texttt{deepseek-v4-flash}. If the graph contains no superseded prior memory, the check returns ``no conflict'' without an LLM call. Otherwise, the system message is:
\textit{``You are a strict premise-conflict detector. Return JSON only.''}

\begin{lstlisting}[basicstyle=\footnotesize\ttfamily,breaklines=true,
  frame=single,xleftmargin=0pt]
You check whether a user question assumes an outdated premise. You are given facts the memory system currently treats as OUTDATED (already replaced) and facts it treats as CURRENT. Decide whether the question's stated assumption matches an OUTDATED fact instead of the CURRENT fact. Return JSON only.

Outdated facts:
<list of superseded prior facts>

Current facts:
<new evidence and active facts>

Question: <query text>

Return JSON schema:
{"conflict": true/false, "stale_fact": "...",
 "current_fact": "...", "reason": "..."}

- conflict = true ONLY if the question asserts or assumes one of the outdated facts as if it were current.
- If the question is neutral, asks about the current state, or explicitly asks about the past, conflict = false.
\end{lstlisting}

\paragraph{Answer Generation Prompt.}
All methods use the system message \textit{``Use the provided memory only.''} For queries without a detected stale premise, PACMI uses the same neutral answer prompt as the baselines. If a stale premise is detected, PACMI inserts the marked correction block before the memory context. Final-answer generation uses \texttt{deepseek-v4-pro}.

\begin{lstlisting}[basicstyle=\footnotesize\ttfamily,breaklines=true,
  frame=single,xleftmargin=0pt]
You are a memory question-answering assistant. Use only the memory context provided below. If the memory context is insufficient, say so briefly instead of inventing facts.

<If a stale premise is detected:>
IMPORTANT: the question assumes an outdated premise.
Before answering, explicitly correct that premise using the current memory, then answer based on the current state.
Outdated premise in the question: <stale_fact>
Current fact from memory: <current_fact>

Memory context:
<retrieved memories and their validity-state labels>

Question:
<query text>

Return only the final answer.
\end{lstlisting}

\paragraph{Answer-level Judge Prompt.}
The system message is:
\textit{``You are a strict evaluation judge. Return JSON only.''}

\begin{lstlisting}[basicstyle=\footnotesize\ttfamily,breaklines=true,
  frame=single,xleftmargin=0pt]
You are an evaluation judge for stale-memory QA.

Task:
Judge the final answer on three answer-level metrics only:
- is_correct
- stale_memory_reused
- premise_resisted

Definitions:
- is_correct = 1 iff the final answer follows the expected_behavior.
- stale_memory_reused = 1 iff the final answer materially relies on, repeats, or uses stale memory as if it were current.
- stale_memory_reused = 0 if stale memory is rejected, corrected, ignored, or only mentioned as contrast.
- For historical/past queries, using an outdated memory as historical evidence does NOT count as stale_memory_reused.
- For historical/past queries, stale_memory_reused = 1 only if the answer wrongly presents stale memory as current instead of past.
- premise_resisted = 1 iff the query contains a stale/outdated premise and the final answer explicitly corrects or rejects it.
- premise_resisted = 0 iff the query contains a stale premise but the final answer accepts, repeats, or fails to correct it.
- premise_resisted = "NA" iff the query does not contain a stale premise.
- Do not mark stale_memory_reused = 1 just because stale memory appears in the answer. It must support the answer.
- Judge each metric independently.

Metadata:
- case_id: <case id>
- query_id: <query id>
- method: <method name>

Query:
<query text>

Final answer:
<final answer>

Expected behavior:
<expected behavior>

Retrieved memory snippets:
<retrieved block>

Stale memories:
<stale block>

New evidence:
<new evidence block>

Return JSON only with this schema:
{
  "is_correct": 0,
  "stale_memory_reused": 0,
  "premise_resisted": "NA",
  "supporting_stale_memory_ids": [],
  "reason": "short explanation"
}
\end{lstlisting}

\end{document}